\documentclass{article}

\usepackage{arxiv}

\usepackage{enumitem}
\usepackage{multicol}
\usepackage{multirow}
\usepackage{graphicx}
\graphicspath{ {./images/} }
\usepackage[rightcaption]{sidecap}
\usepackage{cite}
\usepackage{blindtext}
\usepackage{wrapfig}
\usepackage{amsmath}
\usepackage[hang]{footmisc}
\usepackage{lipsum}
\usepackage{amsmath}
\usepackage{comment}
\usepackage[utf8]{inputenc}
\usepackage{graphicx}
\usepackage{rotating}
\usepackage{subcaption}
\usepackage{xcolor}
\usepackage{comment}
\usepackage{longtable}

\newcommand{\eg}{\textit{e.g.}}
\newcommand{\ie}{\textit{i.e.}}

\definecolor{cvprblue}{rgb}{0.21,0.49,0.74}
\usepackage[pagebackref,breaklinks,colorlinks,citecolor=cvprblue]{hyperref}

\title{Deep Learning for Hate Speech Detection: A Comparative Study}

\author{
Jitendra Singh Malik \footnotemark[2] \\
  School of Computer Science\\
  University of Adelaide\\
  South Australia \\
  \texttt{jmjmalik22@gmail.com} \\
   \And
Hezhe Qiao \footnotemark[2]\\
  School of Computing and Information Systems\\
  Singapore Management University\\
  Singapore \\
  \texttt{hezheqiao.2022@phdcs.smu.edu.sg} \\
   \And
 Guansong Pang  \footnotemark[1]\\
  School of Computing and Information Systems\\
  Singapore Management University\\
  Singapore \\
  \texttt{gspang@smu.edu.sg} \\
  \And
 Anton van den Hengel \\
  School of Computer Science\\
  University of Adelaide\\
  South Australia \\
  \texttt{anton.vandenhengel@adelaide.edu.au} \\
}

\begin{document}
\maketitle
\renewcommand{\thefootnote}{\fnsymbol{footnote}}
\footnotetext[2]{These authors contributed equally to this work.} 
\footnotetext[1]{Corresponding author.} 
\renewcommand{\thefootnote}{\arabic{footnote}}
\begin{abstract}
Automated hate speech detection is an important tool in combating the spread of hate speech, particularly in social media. 
 Numerous methods have been developed for the task, including a recent proliferation of deep-learning based approaches. A variety of datasets have also been developed, exemplifying various manifestations of the hate-speech detection problem.
 %A  However, no such comprehensive empirical evaluation has been reported. 
 We present here a large-scale empirical comparison of deep and shallow hate-speech detection methods, mediated through the three most commonly used datasets.  Our goal is to illuminate progress in the area, and identify strengths and weaknesses in the current state-of-the-art.  
%A This work aims to fill this gap by investigating the performance of 14 diverse state-of-the-art shallow and deep hate speech detection methods on three popular benchmarks. We have a thorough evaluation of these methods in terms of four important aspects, including 
We particularly focus our analysis on measures of practical performance, including
detection effectiveness, computational efficiency, capability in using pre-trained models, and domain generalization.
In doing so we aim to provide guidance as to the use of hate-speech detection in practice, quantify the state-of-the-art, and identify future research directions. Code and data are available at \renewcommand\UrlFont{\color{blue}}\url{https://github.com/jmjmalik22/Hate-Speech-Detection}.
\end{abstract}

\section{Introduction}

Social media has experienced incredible growth over the last decade, both in its scale and importance as a form of communication.  The nature of social media means that anyone can post anything they desire, putting forward any position, whether it is enlightening, repugnant or anywhere between. Depending on the forum, such posts can be visible to many millions of people.
Different forums have different definitions of inappropriate content and different processes for identifying it, but the scale of the medium means that automated methods are an important part of this task.  Hate-speech is an important aspect of this inappropriate content.    

Hate-speech is a subjective and complex term with no single definition, however.  Irrespective of the definition of the term or the problem, it is clear that automated methods for detecting hate-speech are necessary in some circumstances.  In such cases it is critical that the methods employed are accurate, effective, and efficient.

A variety of methods have been explored for the hate speech detection task, including traditional classifiers \cite{waseem2016hateful,davidson2017automated,macavaney2019hate,pamungkas2020misogyny,ayo2021probabilistic}, deep learning-based classifiers \cite{agrawal2018deep,bahdanau2014neural,badjatiya2017deep,raza2022fake,yenala2018deep}, or the combination of both approaches \cite{badjatiya2017deep,mossie2020vulnerable, ibrohim2019multi}. Classifiers like support vector machines (SVM), extreme gradient boosting (XGB), and multi-layer perceptrons (MLP) are commonly used in this task, which typically require  vector representations of the text data. Bag of words models are commonly used, together with TF-IDF (term frequency - inverse document frequency) \cite{aizawa2003information,tang2020several}. With the progress in deep learning-based embeddings, tools such as word2vec \cite{mikolov2013distributed}, Glove \cite{pennington2014glove}, FastText \cite{joulin2016fasttext,bojanowski2017enriching}, and transformer-based methods \cite{devlin2018bert, Clark2020ELECTRA} have been applied to obtain more expressive representations. Both traditional and deep classifiers can be applied to these embedding-based representations pre-trained using the representation learning tools. This substantially increases the pool of methods for hate speech detection, resulting in a large set of possible hate speech detection solutions with different applicability in diverse real-world application contexts. 

On the other hand, there have been a number of dataset benchmarks introduced and released for the evaluation of the performance of these methods, such as Davidson \cite{davidson2017automated}, Founta \cite{founta2018large} and Twitter Sentiment Analysis (TSA)\footnote{\url{https://datahack.analyticsvidhya.com/contest/practice-problem-twitter-sentiment-analysis/?utm_source=av_blog&utm_medium=practice_blog_text_classification##DiscussTab}}. These datasets differ largely from each other in terms of the classes of hatred texts (\eg, sexists, racists, abusive, and offensive tweets), data collection and labeling methods, and data distribution, representing the challenges and application demands from different perspectives. 

To provide insightful application guidelines, in this paper we aim to provide a thorough empirical evaluation and comparison of different types of hate speech detection methods on these datasets. Through this evaluation study, we answer the following four key questions in hate speech detection. i) How is the effectiveness of different popular detection models on diverse hate speech datasets? This is important because practitioners who have different application contexts often need to choose from the large pool of detectors.  ii) Are there any specific models that achieve generally more desired performance than the other models (in terms of both detection effectiveness and efficiency)? Both effectiveness and efficiency are crucial since handling those massive online text data in a timely manner requires computationally scalable and accurate detectors. iii) How effective do popular pre-training methods work with detection models? Pre-training methods have been playing a major role in driving the development of many machine learning and natural language processing areas \cite{camacho2018word,tay2020efficient}, including hate speech detection. It is thus important to evaluate the effectiveness of combining different pre-training and hate speech detectors. iv) How is the generalizability of detection models in tackling domain shifts? This question is included because, due to the diversity in the classes of hate speech, ways of expressing hatred texts, and difference across languages, cross-domain hate speech detection has been emerging as one of the most important problems \cite{fortuna2018survey, chakravarthi2022multilingual, schmidt2017survey}. To our best knowledge, there is no such comprehensive empirical evaluation. The most related work are \cite{badjatiya2017deep,corazza2020multilingual}. Badjatiya  et al. \cite{badjatiya2017deep} presents an empirical comparison of multiple classifiers (including both traditional and deep classifiers) for detecting hatred tweets, but it focuses on the detection effectiveness on  on a single benchmark dataset of 16K tweets. Corazza et al. \cite{corazza2020multilingual} exclusively focuses on empirical evaluation of identifying hate speech across different languages. Our work significantly complements these two studies in both of depth and breadth of the empirical evaluation.

To summarize, this work makes the following two major contributions.
\begin{itemize}
    \item This paper presents a large-scale empirical evaluation of hate speech detection methods to provide insights into their detection effectiveness, computational efficiency, capability in using pre-trained models, and domain generalizability, offering important guidelines for their deployment in real-world applications. As far as we know, this is the first work dedicated to performing such a comparative study to investigate these questions.
    \item We perform a comprehensive evaluation study that involves 14 shallow/deep classification-based hate speech detectors, which are empowered by different word representation methods ranging from TF-IDF, Glove-based word embeddings to advanced transformers. To have an in-depth analysis of the performance in diverse application contexts, these detectors are evaluated on three large and publicly available hate speech detection benchmarks that contain different types of hatred tweet classes from different data sources. All codes are made publicly available at GitHub\footnote{\url{https://github.com/jmjmalik22/Hate-Speech-Detection}}.
\end{itemize}

This paper is organized as follows. We present a brief review of hate speech approaches in Section \ref{sec:approach}, followed by our evaluation approach in Section \ref{sec:evaluation_approach} and a series of our empirical evaluation results in Section \ref{result}. Lastly, the work is concluded in Section \ref{sec:conclusions}.

\section{Hate Speech Detection Approaches}\label{sec:approach}

Many methods have been introduced for hate speech detection \cite{fortuna2018survey,yin2021towards, rini2020systematic, poletto2021resources, austin2020classifying}. In this work we categorize them into three groups, including traditional (shallow) classification methods, word embedding-based deep methods, and transformers-based deep methods. We have a brief review of these methods in this section, and perform comprehensive empirical evaluation of them in the next section.

\subsection{Shallow Methods}
By shallow detection methods, we refer to hate speech detectors that use traditional word representation methods to encode words and apply shallow classifiers to perform the detection.
% of the textual data needs to be done manually to present it into the vectors of feature. These features are then fed into the classifiers.
Various types of such feature representations, such as TF-IDF \cite{aizawa2003information,tang2020several} and n-grams \cite{burnap2015cyber, chen2012detecting,nobata2016abusive,davidson2017automated}, have demonstrated good performance, when combined with traditional classification models \cite{schmidt2017survey,burnap2015cyber, greevy2004classifying,kwok2013locate,mehdad2016characters}. 
Additionally, clustering-based word representation methods \cite{warner2012detecting,del2017hate,nobata2016abusive,xiang2012detecting,zhong2016content} have also been a popular method that can present the texts in lower dimensions.
To capture the underlying sentiments in the texts, sentiment lexicon and the embedded polarity's degree are found to be helpful when modeling the texts for hate speech detection \cite{burnap2015cyber,davidson2017automated,del2017hate, burnap2015cyber,gitari2015lexicon,nobata2016abusive}. Other semantics, such as part of speech and other relevant linguistic features \cite{burnap2015cyber,chen2012detecting,davidson2017automated,gitari2015lexicon,zhong2016content}, as well as word dependency (\eg, `we versus them') \cite{zhong2016content,chen2012detecting} can also be important to have more accurate detection of offensive texts.

% When we talk about the classifiers, it can be stated that
In terms of classification models, support vector machines (SVM) is one of the most popular methods used in hate speech detection \cite{badjatiya2017deep,chen2012detecting,davidson2017automated,greevy2004classifying,mehdad2016characters,warner2012detecting}. Other popular classifiers for this task include naive Bayes \cite{chen2012detecting,davidson2017automated,kwok2013locate,mehdad2016characters}, logistic regression \cite{davidson2017automated,kwok2013locate,waseem2016hateful}, random forest \cite{davidson2017automated}, and gradient boosting decision tree models \cite{saroj2019irlab}.

\subsection{Deep Learning Methods}

Deep learning methods refer to deep neural network-based hate speech detectors.  
% to recognize the features representation from the data provided to it (i.e. input data), which is passed to multi layers present in the network for the classification of the text. It can be said that, 
The input data to these neural networks can be in any form of feature encoding, including traditional methods like TF-IDF and recently emerged word embedding or pre-training methods. The latter approach is generally more effective than the former approach, because it helps avoid traditional feature engineering or feature construction methods. It instead learns feature representations from the presented texts.
Some popular deep neural network architectures include convolutional neural networks (CNN), long short term memory (LSTM) and bi-directional LSTM (Bi-LSTM) \cite{badjatiya2017deep,gamback2017using,del2017hate,park2017one}. 
In hate speech detection, CNN models learn compositional features of words or characters
\cite{badjatiya2017deep,park2017one,gamback2017using}, where LSTM models are used for learning the words that have a long-range dependency of the characters \cite{badjatiya2017deep,del2017hate}.

\subsubsection{Word Embeddings-based Methods}

Word embedding techniques leverage distributed representations of words to learn their vectorized representations to enable downstream text mining tasks \cite{mikolov2013distributed,le2014distributed,pennington2014glove,joulin2016fasttext}. 
% s can be defined as the representation of word in corpus.
The resulting embeddings allow the words with similar meaning to have similar representations in a vector space. 
% They can be defined as belonging to the class of technique where every single words is made to represent by a vector in defined vector space. Every word belong to one vector and all the values are such that it represents a part of neural network, and thereby these are lumped into the areas of the deep learning. The function depends upon creating a dense distributed vector of every word. The vector can have hundreds or millions of dimension to represent the word in sparse.
There have been many word embeddings methods introduced over the years, such as word2vec \cite{mikolov2013distributed}, Glove \cite{pennington2014glove}, and FastText \cite{joulin2016fasttext,bojanowski2017enriching}.
% are one of the many word embeddings which can be used to create vector representation from the text corpus. 
One key intuition behind these models is that the word-word co-occurrence probabilities have the potential for encoding some form of semantic meaning between the words. Readers are referred to \cite{camacho2018word} for detailed review of these embedding methods.

% \noindent \textbf{Relevant Studies}
Word embedding models have been widely used to enable hate speech detection and other relevant tasks, such as sentiment analysis, in a large number of studies \cite{arango2019hate,sharma2017vector,ni2020sentiment,mossie2020vulnerable,mossie2020vulnerable,gamback2017using,poulston2017using,zhang2018detecting,jha2017does,waseem2016hateful}. The learned word embeddings can be combined with traditional classifiers \cite{zhang2018detecting,badjatiya2017deep}, or deep neural network-based classifiers, such as recurrent neural networks (RNN) \cite{arango2019hate,sharma2017vector}, gated recurrent units (GRU) \cite{mossie2020vulnerable}, LSTM \cite{ni2020sentiment}, and CNN \cite{gamback2017using}. The word embedding models help capture semantic and syntactic word relations for detecting hatred tweets, empowering impressive detection effectiveness on different datasets.
% Many researches indicates the use of Glove with the different models like SVM, XGB, Random Forest to give state of art result. Glove can also be fed into RNN by first creating the embedding matrix learned of text corpus by training on pre-trained model.\cite{arango2019hate,8324059}

\subsubsection{Transformers-based Methods}
% \noindent \textbf{Relevant Studies}. 

The above models rely on the usage of LSTM, Bi-LSTM, and CNN, along with the combination of Glove or other classical word embedding technique, which obtain promising performance but they are often not as accurate as the modern transformers-based embedding techniques, such as Small BERT \cite{turc2019well}, BERT \cite{devlin2018bert}, ELECTRA \cite{Clark2020ELECTRA}, and AlBERT \cite{turc2019well}. Readers can refer to \cite{tay2020efficient} for detailed introduction of the transformer models. The transformers can be effectively combined with CNN, LSTM, multi-layer perceptrons (MLP), or Bi-LSTM \cite{mathew2020hatexplain,corazza2020multilingual,awal2021angrybert}, and they enable remarkable performance across hate speech detection datasets in different languages, such as datasets in French, English and Arabic \cite{mathew2020hatexplain}, in Italian, English, Korean and German \cite{corazza2020multilingual, mollas2022ethos, lee2022k}, or in English, Hindi, and German \cite{roy2021leveraging}.

\section{Our Approach for Comparative Study}\label{sec:evaluation_approach}

We aim at answering the following four key questions in hate speech detection through our comparative study.
% through large-scale empirical evaluation and comparison:
\begin{itemize}
    \item Q1: How is the effectiveness of different popular detection models on different hate speech datasets?
    \item Q2: Are there any specific models that achieve generally more desired performance than the other models (in terms of both detection effectiveness and efficiency)?
    \item Q3: How effective do popular pre-training methods work with detection models?
    \item Q4: How is the generalizability of detection models in tackling domain shifts?
    % \item Q5: Does class-imbalanced learning help improve the detection performance?
\end{itemize}

To this end, we perform large-scale empirical evaluation of a large set of shallow and deep methods on three publicly available popular dataset benchmarks.

\subsection{Datasets}\label{subsec:datasets}

Three widely-used datasets, including \textbf{Davidson} \cite{davidson2017automated}, \textbf{Founta} \cite{founta2018large} and Twitter Sentiment Analysis (\textbf{TSA})\footnote{\url{https://datahack.analyticsvidhya.com/contest/practice-problem-twitter-sentiment-analysis/?utm_source=av_blog&utm_medium=practice_blog_text_classification##DiscussTab}}, are adopted in our experiments. 
% One of the important feature for supervised learning is the presence of data which is properly labeled. Thereby, it is utterly important to focus on how the data we are going to use for the study is produced.
Note that to address privacy concerns, the data only constitute of text sentences and the category they belong to. The details of the users are removed according to policy of different websites like Twitter\footnote{https://developer.twitter.com/en/developer-terms/agreement-and-policy}. 
% Now we will provide the information on the dataset we are using in our study.We have used and combined different type of dataset here in our approach. 
An introduction to each of the datasets is presented below, with its key statistics summarized in Table \ref{table:1}.

\textbf{Davidson} \cite{davidson2017automated} is collected starting with the hate speech lexicon that contains different words and phrases identified by various internet users as hate speech.
% , and compiled by Hatebase.org. 
By using the Twitter API, its authors then search for tweets containing terms from the lexicon, resulting in a set of 85.4 million tweets from 33,458 Twitter users. From this corpus, a random set of 25k tweets is sampled and then manually coded by CrowdFlower (CF) workers. Workers are asked to label each tweet as one of three categories \textit{Offensive, Hate, Neither hate speech no offensive}.

\begin{table}[h!]
\centering

\caption{Datasets and their key statistics}
\label{table:1}

\begin{tabular}{ lll } 
\hline
\textbf{Dataset} & \textbf{Class and Statistics}   \\
\hline
\hline
\multirow{4}{5em}{Davidson} & Offensive – 19,190 (77.4$\%$) \\ 
& Hate -  1,430 (5.8 $\%$)  \\ 
& Neither - 4,163 (16.80$\%$) \\
& Total $\sim 25k$  \\ 
\hline

\multirow{4}{5em}{Founta} & None - 53851(53.8 $\%$)  \\ 
& Hate - 4,965 (4.96 $\%$)  \\ 
& Abusive - 27,150 (27.15 $\%$)  \\ 
& Spam - 14,030 ($14.03\%$)  \\ 
& Total $\sim100k$ \\ 

\hline

\multirow{3}{5em}{TSA} & Not Racist/Sexist - 29,720 (92.99$\%$)  \\ 
& Racist/Sexist – 2,242 (7.01 $\%$)  \\
& Total $\sim 32k $ \\

\hline
\end{tabular}

\end{table}

\textbf{Founta} is recently published in \cite{founta2018large}, which contains around 100k human annotated tweets with about 27\% being abusive, 14\% being spam, and 5\% being hate speech. The data was completed in different rounds. In the first round, annotators divided the tweets into three classes \textit{normal, spam and inappropriate}. Thereafter, annotators were explained to further reclassify the tweets of the category \textit{inappropriate}. The final version of the dataset includes four classes -- \textit{normal, spam, hate, and abusive.}
 
\textbf{TSA}
% - the given dataset for detecting hatred tweets 
is released by Analytic Vidhya\footnote{\url{https://datahack.analyticsvidhya.com/contest/practice-problem-twitter-sentiment-analysis/?utm_source=av_blog&utm_medium=practice_blog_text_classification##DiscussTab}} on the competition - Twitter Sentiment Analysis\footnote{\url{https://www.kaggle.com/dv1453/twitter-sentiment-analysis-analytics-vidya}}. The dataset is composed by about 32K tweets,
% The dataset is provided in the form of a csv file with each line storing a tweet id, its label and the tweet. 
and contains only two classes namely \textit{ `Not Racist/Sexist' and `Racist/Sexist' }

\subsection{Detection Models}\label{Model}

We consider three types of detection methods base on how they perform word embeddings, as the feature representation is the key to hate speech detection.

\textbf{Traditional Classifiers}. This type of methods comprises of applying shallow classifiers -- Support Vector Machine (SVM), extreme gradient boosting (XGB), and multi-layer perceptrons (MLP) -- on top of the TF-IDF-based word representations. Particularly, the TF-IDF algorithm is a statistical method to measure the relevancy of a word in a document in a collection of the documents. The TF-IDF-based vector embeddings are then used with one of the traditional classifiers \cite{badjatiya2017deep}.

\textbf{Deep Models with Glove Embeddings}. This approach uses the Glove-based embedding model with deep classifiers using CNN, MLP or Bi-LSTM-based network architecture. The classifier is trained with cross-entropy loss.

\textbf{Deep Models with Transformers-based Pretraining}. This approach is focused upon the transformer models, including Small BERT \cite{turc2019well}, BERT \cite{devlin2018bert}, ELECTRA \cite{Clark2020ELECTRA}, and AlBERT \cite{turc2019well}, with each in combination with CNN and MLP separately.  
% in these we have used the transformers-based embeddings to try to increase our accuracy and obtain better results.
BERT and other Transformer encoder architectures have been successful on a variety of tasks in NLP (natural language processing). They compute vector-space representations of natural language that are suitable for use in deep learning models. The BERT family of models uses the Transformer encoder architecture to process each token of input text in the full context of all tokens before and after. The classifier is also trained with cross-entropy loss.
% , hence the name: Bidirectional Encoder Representations from Transformers.

All these together result in 14 popular/state-of-the-art models in this study, including three TF-IDF-based traditional classifiers, three Glove-based models, and eight transformer-based models.

\subsection{Implementation Details}

We present the implementation details from three consecutive steps, including embedding, detection model, and optimization.

\textbf{Embedding}. A sentence of length $n$ can be represented as $ w_1, w_2....w_n$ where each word can be represented as a real valued vector. For word embedding, we deploy TF-IDF, the pre-trained Glove embedding of Tweets, or transformers-based embedding. In TF-IDF, each tweet is represented by a vector with a dimensionality size of the dictionary (a collection of unique words across all training tweets) size, in which each entry is denoted by the multiplication of the frequency (TF) and inverse document frequency (IDF) of a specific term (word). This simple embedding method is used in three traditional classifiers: SVM, XGB and MLP. Note that due to extensive computation it takes, we limited the dictionary size to maximum 10k in our experiments.
% (as discussed in section \ref{Model}.)

Glove is generally a log-bilinear model with a weighted least-squares objective. The main intuition underlying the model is the simple observation that the common word-word co-occurrence patterns have the potential for encoding some form of semantic. The advantage of this model is that we can leverage massive datasets with billions of words that one may not have access to, to capture word meanings in a statistically robust manner. In our implementation, 
% the Glove-based embedding layer processes a fixed size sequence of words. Each word is represented as a real-valued vector, also known as word embeddings. In this work, we tested 
we use pre-trained Glove word vectors with a dimensionality size of 100. This Glove-based embedding layer is subsequently connected to CNN, MLP, or Bi-LSTM to train deep classifiers.
% that are trained from Tweets.

The BERT family of models have similar advantages as Glove, but can produce more meaningful word embeddings. In our experiment, we utilize the pre-trained BERT models available on TensorFlow Hub\footnote{https://www.tensorflow.org/hub}, which allows us to easily integrate BERT in our implementation. Four different types of transformer-based embedding models are used, including 
\begin{itemize}
    \item BERT\footnote{bert\_en\_uncased\_L-12\_H-768\_A-12/3 at https://tfhub.dev/google/collections/bert/1}, which is pre-trained on a large corpus of text, and then fine-tuned for specific tasks  
    % with Talking-Heads Attention and Gated GELU [base, large] has two improvements to the core of the Transformer architecture.
    \cite{devlin2018bert}.
    \item Small BERT\footnote{small\_bert/bert\_en\_uncased\_L-2\_H-12\_A-2/1 at \url{https://tfhub.dev/google/collections/bert/1}}, which is an instance of the original BERT architecture with a smaller number L of layers (\ie, residual blocks) combined with a smaller hidden size H and a matching smaller number A of attention heads \cite{turc2019well}. Small BERTs have the same general architecture but fewer and/or smaller Transformer blocks, enabling a good trade-off between speed, size and quality.
    \item ALBERT\footnote{albert\_en\_base/2 at \url{https://tfhub.dev/google/albert_base/3}}, which is ``A Lite" version of BERT with greatly reduced number of parameters \cite{turc2019well}. ALBERT computes dense vector representations for natural language by using a deep neural network with the transformer architecture. 
    % There are four different sizes of AlBERT that reduces model size (but not computation time) by sharing parameters between layers.It was originally published in study \cite{}.
    \item ELECTRA\footnote{google/electra\_small/2 at \url{https://tfhub.dev/google/collections/electra/1}}, which is a BERT-like model that is pre-trained as a discriminator in a set-up resembling a generative adversarial network (GAN) \cite{Clark2020ELECTRA}.

\end{itemize}

Similar to \cite{vijayaraghavan2021interpretable, awal2021angrybert,mathew2020hatexplain,awal2021angrybert}, the BER models are subsequently connected to a CNN/MLP architecture to train the deep detectors.

\textbf{Network Architecture}.
In this section we introduce the implementation of each model in our study. More specifically,
% The complete structure for each type of model has been explained in terms of layers, optimizer and parameters. Our architecture consists of several layers, starting with an embedding layer, with next layer according to the model as described below:
we use SVM with linear kernel and XGB with default parameters as in the Scikit-learn library\footnote{https://scikit-learn.org/}.
% with the TF-IDF output been directly fed into this models
% . TF-IDF embedding were made to be calculated by first training the data on the the whole corpus and then obtaining the embeddings using sklearn library. 

\begin{figure}
\centering
  \includegraphics[width=1.0\textwidth, angle=0]{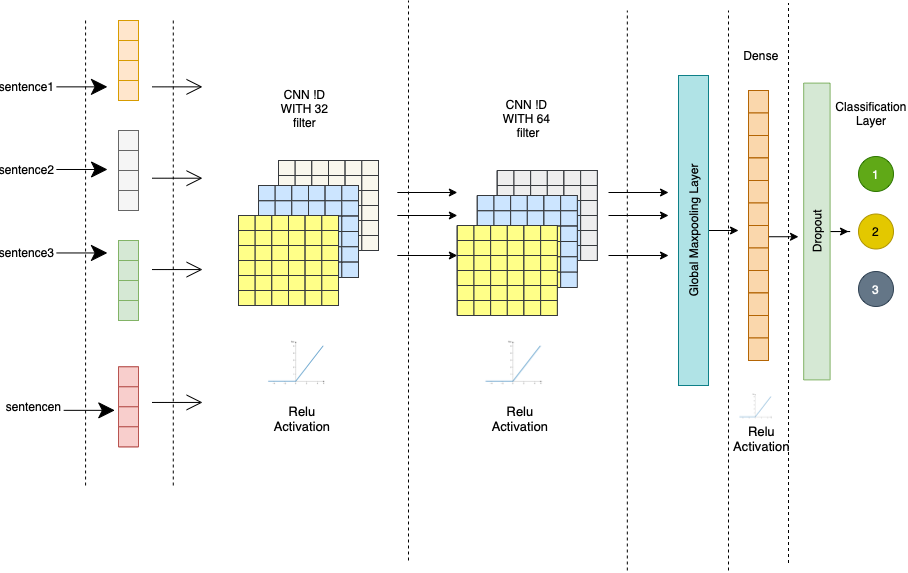}
   \caption{CNN Architecture used in this study}
  \label{cnn}
\end{figure}

% SVM model was provided with the manual weights as given in the section \ref{cweight} whereas the XGB models was made to train using the automated weights calculated by default by model itself.
For MLP, its architecture consists of
% of MLP has been kept simple with a transformer embedding layer,
one dense layer, a dropout layer with the probability of 0.1, and one classification layer. For the CNN architecture, as shown in Figure \ref{cnn}, it consists of two CNN layer with filter size of 32 and 64, respectively. The output from the last CNN layers is fed to a maxpooling layer, followed by a dense layer with 256 neurons and a dropout layer with the probability of 0.1, before feeding to the last output layer. 
% and dense layer with 2 neurons to classify the class.

\begin{figure}
   \centering
  \includegraphics[width=1.0\textwidth, angle=0]{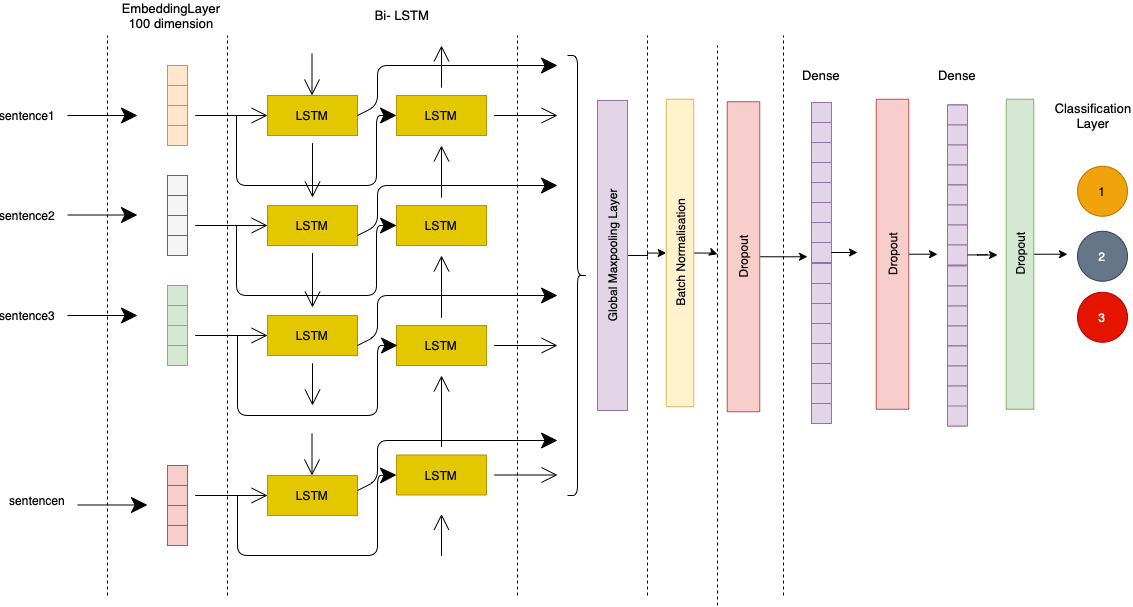}
  
  \caption{Bi-LSTM Network Architecture used in this study}
  \label{bilstm}
\end{figure}

In Bi-directional LSTM that is used in the Glove-based model, it is composed of Glove Embedding Layer (dimension=100), followed by Bi-LSTM with recurrent dropout of a probability equal to 0.2, a Global Max pooling layer of one dimension, a Batch Normalistaion layer. It is then followed by the combination of dropout layer with a probability of 0.5 and a dense layer with the relu activation, and lastly
% except the last dense layer consists of number of neurons equal to two or three (according to number of classes in dataset) with 
a softmax classification layer to predict the the classes. The complete architecture is shown in Figure \ref{bilstm}.

\textbf{Optimization}. 
SVM and XGB are trained with the recommended settings in Scikit-learn. The neural network-based classifiers are trained with a batch size of 32, 128 or 256 for 10-25 epochs. 
% The dropout probability is set to 0.1-0.2 for all layers. 
Adam optimizer is used with a learning rate of 2e-5. As input for transformer embedding, we tokenize each tweet with the BERT tokenizer. It contains invalid characters removal, punctuation splitting, and lowercasing the words. Based on the original BERT, we split words to sub word units using WordPiece tokenization. As tweets are short texts, we set the maximum sequence length to 64 and in any shorter or longer length case it is padded with zero values or truncated to the maximum length. We use the cross entropy loss function and the Adam optimizer to optimize the models.

All experiments are performed using a Google co-laboratory tool which is a free research tool with a Tesla K80 GPU and 12G RAM and Kaggle environment.

\subsection{Performance Evaluation Metrics}

\noindent \textbf{Per-class Metric}. 
The results of different methods are measured based on 
% will be classified and compared to each other based on three factors which are obtained from the classification report of sklearn package. Study \cite{chicco2020advantages} defines the explanation of the
Precision, Recall and $F_1$ score \cite{chicco2020advantages} defined as below:

\begin{itemize}
   \item \textbf{Precision} $p_i$ - Precision is the fraction of true positive examples among the examples that the model classified as positive for a specific class. 
   
   \item \textbf{Recall} $r_i$- Recall is the fraction of examples classified as positive among the total number of positive examples per class.
   
    \item \textbf{$F_1$ score} - $F_1$ score can be defined as the harmonic mean of precision and recall: $F_1 = 2 \frac{p_i . r_i }{ (p_i + r_i)}$.

\end{itemize}

%\noindent \textbf{Micro Metric}.
%the micro metric in simple words can easily be defined as the 

\noindent \textbf{Evaluation Metric}. For the overall performance, we consider both of macro $F_1$ score and weighted average $F_1$ score of the classification results, as the data is highly imbalanced in the three data sets in Table \ref{table:1}. 
% The main reason of our focus on both measures is because the considering only the one will provide inadequate details of model accuracy which can be explained as such :
\begin{itemize}
    \item Macro $F_1$ score calculates the mean of the $F_1$ scores across all classes: $
     Macro\;F_1 = \frac{1}{C} \sum_{i=1}^{i=C} F_{1i} $, where $C$ is the number of classes and $F_{1i}$ is the $F_1$ score of class $i$. It considers all classes equally.
     
     \item Weighted $F_1$ score, on the other hand, considers the individual weight of each class  \cite{kapil2020deep}, and is defined as $Weighted\;F_1 = \frac{1}{N} \sum_{i=1}^{i=N} F_{1i} N_i$, where $N = \sum_{i=1}^{i=N} N_i$ and $N_i$ is the sample size of class $i$.
\end{itemize}

\section{Experimental Results \& Analysis \label{result}}

\subsection{Q1: Effectiveness of Different Models on Popular Benchmarks}
% We will focus on each dataset separately to find the best embedding which is outperforming the others. We are providing macro and weighted average both for describing the result.
The macro and weighted average results of all 14 hate speech detection models on three popular benchmarks are shown in Table \ref{table:2}. Below we discuss the results on each dataset.

\textbf{Results on the Davidson Dataset\label{wzls}}. 
In terms of macro $F_1$ score, it can be observed that Glove embedding is better than TF-IDF in predicting hate speech more accurately. The results obtained by Glove embedding with CNN and Bi-LSTM are similar to the highest score obtained by TF-IDF embeddings with XGB. This high score was almost touched twice by the glove embedding. The TF-IDF-based MLP and SVM models obtain similarly good performance as the Glove-based MLP model.
In terms of weighted average $F_1$ score, it can be found that XGB with TF-IDF produced very competitive results, which is more effective than all Glove-based models. 
% It produced second highest $F_1$ score recorded among all the models including the transformer embeddings.

Compared to the TF-IDF-based and Glove-based models, the transformers-based models achieve a significant increase in both metrics, achieving the best macro $F_1$ score of 0.76 and the best weighted average $F_1$ score of 0.91. More specifically, in terms of macro $F_1$, all the transformers models outperform every single model in combination with TF-IDF or Glove by a large margin. 
% It is observed that usage of Transformers encoding with neural network increases the $F_1$ score drastically. 
These results demonstrate that the transformers are able to perform better on both large and small classes, especially on the small classes (\ie, the hate class), when compared to Glove and TF-IDF. Small BERT is less effective than TF-IDF-based XGB in weighted average $F_1$ score, indicating that Small BERT underperforms XGB on the large class (\ie, the offensive class). Overall, BERT-based CNN and ELECTRA-based MLP turn out to be the best performers in this dataset.
% with second spot being taken by Small BERT and ELECTRA with CNN combination.
% The same trend can be spotted with weighted average, where the transformers embedding proved to be very successful. Six models out of eight models touched the $F_1$ score of .90.

\begin{table}[htbp]
\caption{Macro and weighted average $F_1$ score performance. The best performance is highlighted in red, while the second best is in blue.}
\label{table:2}
\centering
%%\resizebox{\textwidth}{!}
% \fontsize{8}{12}\selectfont

\begin{tabular}{lllll|lll}
\hline

\multirow{2}{3em}{\textbf{Embedding}} &\multirow{2}{5em}{\textbf{Model}} & \multicolumn{3}{ c }{\textbf{Macro}} & \multicolumn{3}{ c }{\textbf{Weighted Avg.}}  \\ 

% \cline{3-8}

&& \textbf{P} & \textbf{R} & \textbf{$F_1$} & \textbf{P}&\textbf{R}&\textbf{$F_1$}  \\
\hline
\hline
%----------------------- WZLS

\multicolumn{8}{ c }{\textbf{Results on the Davidson Dataset}}\\
\hline

\multirow{3}{4em}{TF-IDF} 

&TF-IDF + SVM & 0.69&0.64&0.66&0.86&0.87&0.86  \\
&TF-IDF + XGB & 0.74&0.69&0.70&0.89&0.90&\textbf{\color{blue}0.90} \\
&TF- IDF + MLP &  0.68&0.63&0.66&0.85&0.86&0.85\\

\hline
%%\cline{3-8}
\multirow{3}{3em}{Glove}

&Glove + CNN &0.66&0.73&0.69&0.88&0.85&0.86
\\
&Glove + MLP &0.72&0.61&0.65&0.85&0.86&0.85
\\
&Glove + Bi-LSTM & 0.67&0.75&0.69&0.88&0.84&0.86\\
\hline
        
%%\cline{3-8}
\multirow{8}{6em}{Transformers} &

Small BERT + CNN & 0.72 & 0.82 & \textbf{\color{blue}0.75} & 0.91 & 0.85 & 0.87 \\
&Small BERT + MLP & 0.71 & 0.81 & 0.74 & 0.91 & 0.85 & 0.87 \\
&BERT + CNN & 0.78 & 0.75 & \textbf{\color{red}0.76} & 0.91 & 0.91 & \textbf{\color{red}0.91} \\
&BERT + MLP & 0.75 & 0.72 & 0.74 & 0.90 & 0.90 & \textbf{\color{blue}0.90} \\
&Al-BERT + CNN & 0.76 & 0.69 & 0.72 & 0.89 & 0.90 & \textbf{\color{blue}0.90} \\
&Al-BERT + MLP & 0.77 & 0.72 & 0.74 & 0.90 & 0.91 & \textbf{\color{blue}0.90} \\
&ELECTRA + CNN & 0.75 & 0.75 & \textbf{\color{blue}0.75} & 0.91 & 0.91 & \textbf{\color{red}0.91} \\
&ELECTRA + MLP & 0.75 & 0.76 & \textbf{\color{red}0.76} & 0.91 & 0.91 & \textbf{\color{red}0.91} \\ 
\hline

%--------------FOunta

\multicolumn{8}{ c }{\textbf{Results on the Founta Dataset}}\\
\hline\hline
\multirow{3}{4em}{TF-IDF}

&TF-IDF + SVM& 0.63&0.71&0.64&0.80&0.73&0.75
\\
&TF- IDF + XGB&0.74&0.59&0.62&0.79&0.80&\textbf{\color{blue}0.78}
\\
&TF- IDF + MLP & 0.64&0.61&0.62&0.75&0.76&0.76\\

\hline
%%\cline{3-8}
\multirow{3}{3em}{Glove}

&Glove + CNN&0.58&0.54&0.52&0.73&0.70&0.69
\\
&Glove + MLP&0.59&0.61&0.59&0.74&0.72&0.73
\\
&Glove + Bi-LSTM&0.63&0.65&0.63&0.77&0.75&0.76
\\
\hline
        
%%\cline{3-8}
\multirow{8}{6em}{Transformers} &

Small BERT + CNN&0.63&0.71&0.65&0.80&0.73&0.75\\
&Small BERT + MLP&0.64&0.71&0.66&0.79&0.74&0.76\\
&BERT + CNN&0.68&0.67&\textbf{\color{blue}0.67}&0.79&0.79&\textbf{\color{red}0.79}\\
&BERT + MLP&0.68&0.67&\textbf{\color{red}0.68}&0.79&0.79&\textbf{\color{red}0.79}\\
&Al-BERT + CNN&0.68&0.68&\textbf{\color{red}0.68}&0.79&0.79&\textbf{\color{red}0.79}\\
&Al-BERT + MLP&0.68&0.67&\textbf{\color{blue}0.67}&0.79&0.79&\textbf{\color{red}0.79}\\
&ELECTRA + CNN&0.65&0.73&0.66&0.80&0.73&0.75\\
&ELECTRA + MLP&0.66&0.73&\textbf{\color{red}0.68}&0.81&0.76&\textbf{\color{blue}0.78} \\
\hline\hline

%----------------------TSA 

\multicolumn{8}{ c }{\textbf{Results on the TSA Dataset}}\\
\hline

\multirow{3}{4em}{TF-IDF} 

&TF-IDF + SVM&0.85&0.72&0.77&0.91&0.92&0.91 \\
&TF- IDF + XGB&0.70&0.99&0.76&0.98&0.95&0.96 \\
&TF- IDF + MLP&0.84&0.78&0.81&0.95&0.95&0.95\\

\hline
%%\cline{3-8}
\multirow{3}{3em}{Glove}  
&Glove + CNN&0.74&0.81&0.77&0.94&0.93&0.94 \\
&Glove + MLP&0.74&0.81&0.77&0.94&0.93&0.94 \\
&Glove + Bi-LSTM &0.78&0.84&0.80&0.95&0.94&0.95 \\
\hline
        
%%\cline{3-8}
\multirow{8}{6em}{Transformers} &

Small BERT + CNN&0.84&0.83&0.84&0.96&0.96&0.96 \\
&Small BERT + MLP&0.82&0.82&0.82&0.95&0.95&0.95\\
&BERT + CNN&0.93&0.88&\textbf{\color{red}0.90}&0.98&0.98&\textbf{\color{red}0.98}\\
&BERT + MLP&0.93&0.87&\textbf{\color{red}0.90}&0.97&0.98&\textbf{\color{blue}0.97}\\
&Al-BERT + CNN&0.91&0.90&\textbf{\color{red}0.90}&0.97&0.97&\textbf{\color{blue}0.97}\\
&Al-BERT + MLP&0.86&0.84&0.85&0.96&0.96&0.96\\
&ELECTRA + CNN&0.90&0.87&\textbf{\color{blue}0.89}&0.97&0.97&\textbf{\color{blue}0.97}\\
&ELECTRA + MLP&0.90&0.87&\textbf{\color{blue}0.89}&0.97&0.97&\textbf{\color{blue}0.97}\\
\hline
%%\cline{3-8}
\end{tabular}

\end{table}

\textbf{Results on the Founta Dataset\label{founta}}. The Founta dataset is more diversified than the other two datasets. Therefore, the performance results within the models vary more compared to the other two datasets. Interestingly, the TF-IDF-based models are generally more effective than the Glove-based models. In terms of macro average, TF-IDF-based SVM obtains very good results with a Macro $F_1$ score of 0.64, outperforming all Glove-based models and performing comparably well to some transformers-based models. Nevertheless, transformer embeddings in this case enable the most superior results to other embeddings. BERT, ALBERT, and ELECTRA (with CNN) prove to be much more successful than other models. 

% In terms of weighted average it was found the TF-IDF in combination of
Similar to the results on Davidson, the TF-IDF-based XGB outperforms the other combination with SVM and MLP in weighted average $F_1$, obtaining an $F_1$ score of 0.78 which also achieves the second highest weighted $F_1$ score among all the models.  
% When compared in terms of weighted average, it was found that the results with TF-IDF embeddings were quite superior to Glove embeddings, majorly in combination with XGB reaching the $F_1$ score of 0.78. When comparing to other TF-IDF combination with Glove. The later combination with Bi-LSTM reached the score of 0.76 which was similar to former combination with MLP and one percent higher with SVM.
Transformers-based models come with the best results. Particularly, the BERT and Al-BERT models dominate the other models with the weighted $F_1$ score reaching to 0.79 in both CNN and MLP, followed by ELECTRA at 0.78 with MLP.

% The summary of the results are described in table \ref{table:2}.

\textbf{Results on the TSA Dataset\label{av}}. The results of most methods on the TSA dataset are much better than the result on the other two datasets, which may be due to the reason that fewer classes and more training samples are presented for each class (see Table \ref{table:1}). 
The results in terms of the macro $F_1$ show similar performance as in previous datasets, with transformers embeddings performing far better than the other two embeddings, achieving a macro $F_1$ score as high as 0.90. 
% Compared to other models, it is observed here that, 
On this dataset, the TF-IDF-based XGB is not as successful as the TF-IDF-based MLP that achieves the macro $F_1$ score of 0.81. The TF-IDF-based MLP also outperforms all Glove-based models.
% Usage of MLP gave better result making the macro $F_1$ score reaching to 0.81 with SVM and XGB reaching to 0.77 and 0.76 with just a litter difference with each other.
% Glove embedding gave slightly better results compared to TF-IDF. 
% In glove CNN combination was a little better than others with $F!_1$ score of 0.80 with CNN and MLP standing at 0.77 each.
In transformer embeddings, BERT (with CNN/MLP) and Al-BERT (with CNN) perform far better than other embeddings with the $F_1$ score of 0.90, followed by ELECTRA, with an $F_1$ score of 0.89 (CNN, MLP).

The weighted average score results are very different from the macro average. The TF-IDF-based MLP and XGB obtain the best results with the weighted $F_1$ score of 0.95 and 0.96, respectively. In terms of Glove embedding, the BI-LSTM model performs similarly well to the TF-IDF-based MLP, and outperforms the CNN/MLP model. The transformers-based models are more effective, with the results obtained by BERT, Al-BERT, and ELECTRA achieving an $F_1$ score of 0.97/0.98. Small BERT and ALBERT (with ML) are slightly less effective compared to other transformers-based models.

\subsection{Q2: Superiority of Specific Models in both Effectiveness and Efficiency\label{accuracy}}

\textbf{Effectiveness}. As shown in Table \ref{table:2}, the Glove and TF-IDF models do not have consistently superior effectiveness over each other. It is interesting that the TF-IDF-based XGB is able to perform better or on par, compared to the results obtained by Glove-based neural network models. Similar to our study, it was also observed in \cite{arango2019hate} that XGB was able to obtain better results compared to Glove-based deep models \cite{agrawal2018deep}. 

% The study \cite{arango2019hate} used the glove embeddings with the XGB whereas our results was based on using the same with the TF-IDF embedding.

The use of Bidirectional LSTM is quite successful in our experiments and obtains good results across three datasets. 
% and only lagged behind the transformers. It can therefore be stated
This indicates that Bi-LSTM is successful in recognizing the underlying relationships in a set of data better than most of the other models except transformers.

In terms of transformers embedding, the use of BERT, ELECTRA, and Al-BERT provide the best results in all three datasets, showing quite consistently superior performance over the competing models. 
The higher performance of the transformer embeddings can also be found in other studies \cite{mozafari2019bert}, in which the base version of BERT was used in combination with CNN and LSTM networks.

\begin{figure}[h]
\centering
    \includegraphics[width=0.95\textwidth,angle=0]{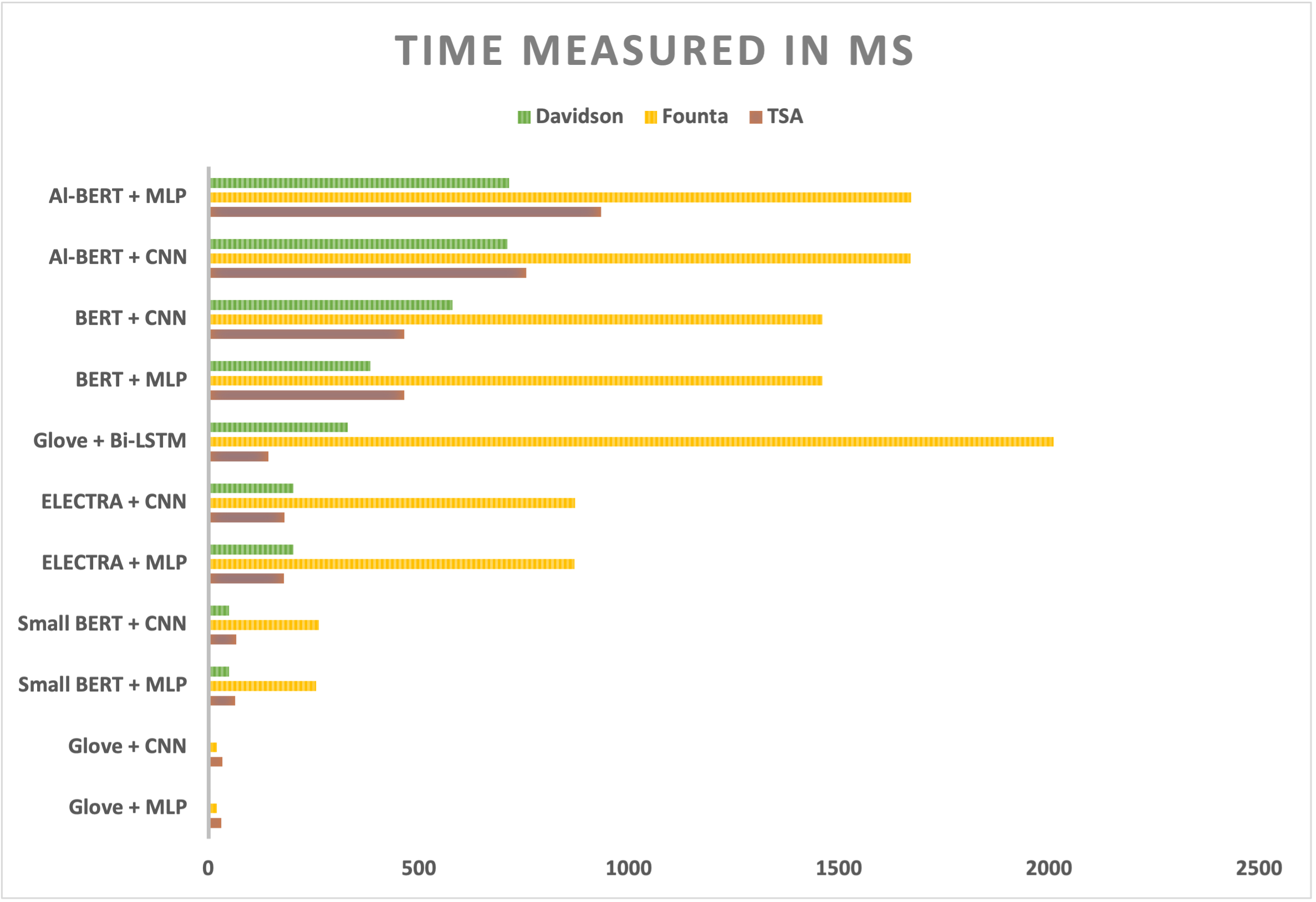}
    \caption{Computational time of the models per epoch measured in ms}
    \label{ct}
\end{figure}

\textbf{Computational Efficiency}. In order to measure the computational time of the models, we consider the running time  to complete one epoch for every model. The results of computational time are shown in Figure \ref{ct}. It is observed that while the effectiveness provided by the transformers is higher than the other embeddings, it does require much more time to train on the dataset. 
% It is due to the fact that the pre-trained transformer-based embeddings are trained on large amount of data compared to embeddings like Glove. 
It can be found that AI-BERT is the most time-consuming model among the transformers, followed by the BERT model in all there datasets we have tested. Note that Bi-LSTM in combination with Glove is significantly more costly than Glove-based CNN and MLP models, which also take a long time to train than ELECTRA and Small-BERT. Small BERT is the most efficient model among all the transformers due to the smaller number of trainable parameters than the other versions of BERT.
% the less amount of dataset it is trained on compared to other transformer embedding. 
% Glove with Bi-LSTM are quite higher in compare to combination of glove with CNN and MLP.

Overall, ELECTRA-based MLP models seem to be the most practical method that can normally achieve the best classification effectiveness while at the same time being sufficiently computation efficient.

% Glove-based CNN and MLP models comes up as the fastest but least accurate compared the transformers. TF-IDF embedding are the quickest among all the models.

\subsection{Q3: Pre-training in Deep Hate Speech Detection Models}

This section discusses how large-scale pre-training helps the subsequent hate speech detection. 
% It has been observed that the pre-trained embeddings were quite successful in our experiments which can be observed by looking at the results of transformers models.
There are two types of pre-training models used in our study: the Glove-based models and transformers-based models. Glove is pre-trained on Twitter with 2B tweets, 27B tokens, a vocabulary of size 1.2M, and 100 dimensions. Transformers are pre-trained on the Wikipedia and BooksCorpus. TF-IDF-based models can be treated as non-pre-trained models. As shown in Table \ref{table:2}, the pre-trained models (\eg, Glove-based MLP and transformers-based MLP) perform generally better than the plain models (\eg, TF-IDF-based MLP) on the three datasets. The transformers-based pre-training is typically much better than the TF-IDF-based MLP models. 

Compared to Glove-based pre-training, different transformers-based pre-training methods perform consistently better across all three datasets in both macro $F_1$ and weighted average $F_1$ measures. This also applies to both CNN and MLP-based classifiers.

\subsection{Q4. Cross-domain Hate Speech Detection \label{cde}}

% \subsubsection{Experimentation Details}

In this section, we investigate the effectiveness of different models in generalizing from one domain to another domain to detect hate speech. The domain difference is mainly due to the source of the datasets. One problem here is that the three datasets have different sets of classes. To tackle this issue, we remove less relevant classes and focus on detecting hatred tweets. The resulting datasets 
% To perform cross domain experiments we have initialized some changes in our existing datasets to form a new modified datasets. The changes 
are shown in Table \ref{cross domain dataset}.  

\begin{table}[h!]
\centering

\caption{Datasets for Cross-domain Experiments}
\label{cross domain dataset}

\begin{tabular}{ ll} 
\hline
% \textbf{Dataset}& 
\textbf{Dataset} & \textbf{Details}   \\
\hline
\hline
%----------- Dataset D1

% \multirow{3}{5em} {\textbf{D1}}  &
\multirow{3}{5em} {Davidson}  
& Hate -  1430 (34.05 $\%$)  \\ 
& Neither - 4163 (65.95$\%$) \\
& Total $\sim 5.5k$  \\

%----------- Dataset D2

\hline
% \multirow{4}{5em} {\textbf{D2}}  & 
\multirow{3}{5em}{Founta} & None - 52885 (62.3 $\%$)  \\ 
& \multirow{1}{29em} {Hate (Combination of Hate - 4965 and Abusive - 27150 (37.7 $\%$) )} \\ 
& Total $\sim85k$ \\ 

\hline

%----------- Dataset D3

% \multirow{4}{5em} {\textbf{D3}} &  

\multirow{3}{5em}{TSA} & 

\multirow{1}{20em} {Neither (Not Racist/Sexist) - 29720 (92.99$\%$)} \\ 
& \multirow{1}{18em}{Hate (Racist/Sexist) – 2242 (7.01$\%$)}  \\
&Total $\sim 32k $ \\

\hline
\end{tabular}

\end{table}

\begin{figure}[t!]
\centering
    \includegraphics[width=0.8\textwidth,angle=0]{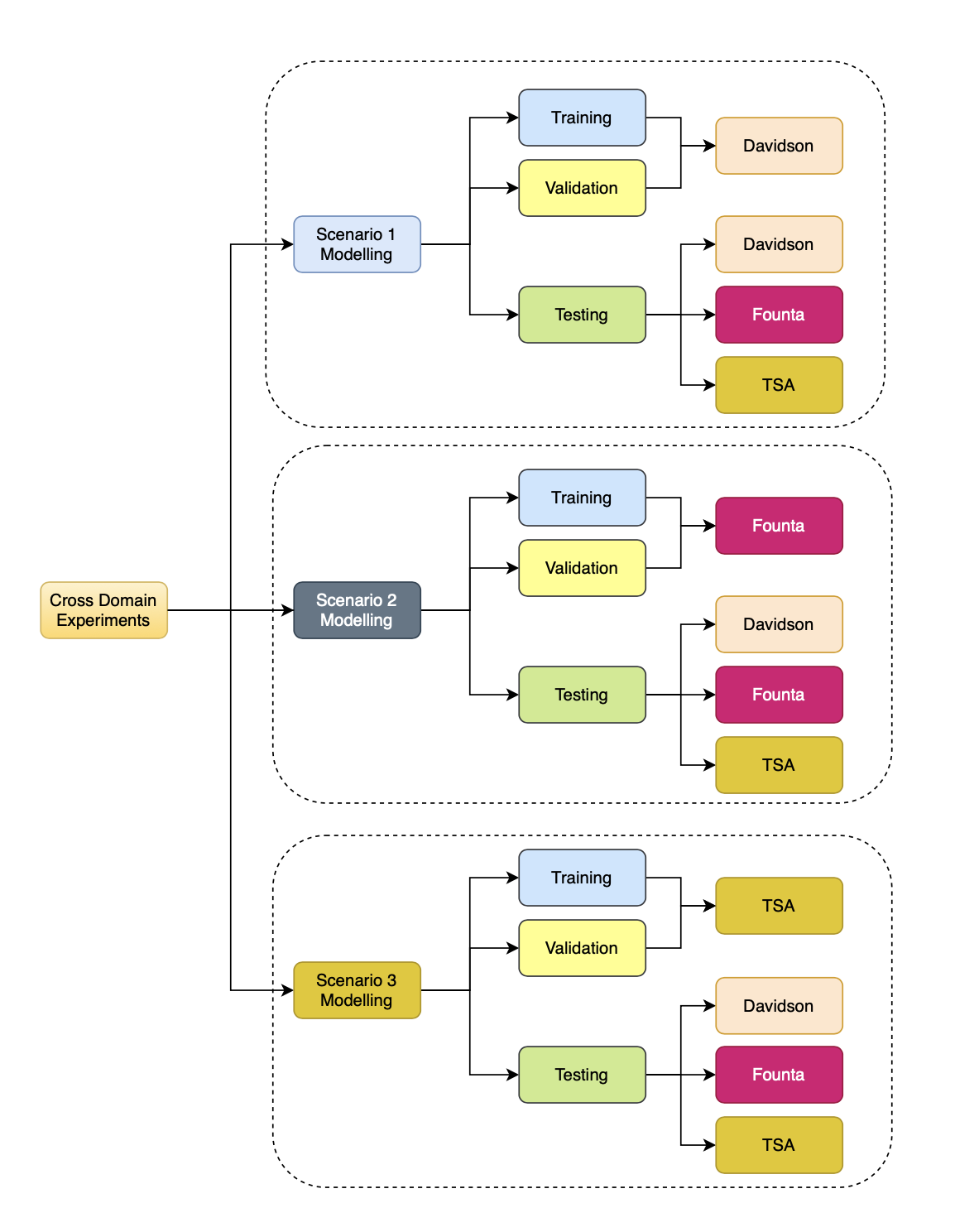}
    \caption{Three scenarios of cross-domain experiments}
    \label{cross-domain}
\end{figure}

We then perform three sets of cross-domain experiments, with each model trained on one of the three datasets and evaluated on the other two datasets, \ie, one dataset is used as the source domain, while the other two datasets are used as the target domain. In addition, we also present the results of the test data of the source domain dataset to serve as baseline performance. The three cross-domain scenarios are illustrated in Figure \ref{cross-domain}. In this experiment, we focus on models that obtain very good performance in Table \ref{table:2} as less effective models in non-cross-domain settings are expected to perform poorly in cross-domain settings.

\textbf{Results in Scenario 1}. Experiment scenario 1 is to train models on Davidson  and test models on the other two datasets. The results are presented in Table \ref{crossd-wzls}. Compared to the results on the test data of the source domain dataset Davidson, the cross-domain performance drops significantly on both Founta and TSA, especially on Founta where all Glove and transformers models obtain $F_1$ score at around 0.5 to 0.6 in both macro $F_1$ and weighted average $F_1$ scores. The performance on TSA gets better for most of the models, particularly on the weighted average results. This indicates the non-hatred tweets in TSA share some common features with that in Davidson, resulting in a good performance on the non-hatred tweet classification and thus a high weighted average $F_1$ score.
% It can be noted here that transformer embeddings were quite successful and provided state of art results.

In the case when the testing is performed on the Davidson dataset, ELECTRA-based CNN models outperform the other models and obtain the best performance with an $F_1$ score of 0.94, followed by that of other transformer embeddings. Glove models are  good at detection but it is still not comparable with transformers embeddings. In terms of weighted average score, the model with the best performance is ELECTRA with CNN, achieving an $F_1$ score of 0.95, followed by other transformed embeddings. Glove-based CNN models provide a score notably equal to Small BERT-based models. Note that the results here are different from that in Table \ref{table:2}, as the datasets are reduced and typically become simpler datasets.

\begin{table}[h!]
\centering
%%\resizebox{\textwidth}{!}
% \fontsize{8}{12}\selectfont

\caption{Scenario 1 - Cross-domain experimental results with the Davidson dataset as the source domain dataset.}
\label{crossd-wzls}

\begin{tabular}{llllllllll}
\hline

\multirow{2}{3em}{\textbf{Embedding}} &\multirow{2}{5em}{\textbf{Model}} & \multicolumn{3}{ c }{\textbf{Macro}} & \multicolumn{3}{ c }{\textbf{Weighted Avg.}}  \\ 

% \cline{3-8}

&& \textbf{P} & \textbf{R} & \textbf{$F_1$} & \textbf{P}&\textbf{R}&\textbf{$F_1$}  \\
\hline
\hline

% \multicolumn{8}{c}{\textbf{Training/ Validation - Davidson}} \\  
% \hline 

\multicolumn{8}{c}{\textbf{Testing - Davidson}} \\ 
\hline

%%\cline{3-8}
\multirow{2}{3em}{Glove} 

& Glove + CNN  & 0.90 & 0.90 &  0.90 & 0.92& 0.92& 0.92          \\

& Glove + MLP   & 0.84         & 0.84         & 0.84          & 0.88          & 0.88          & 0.88          \\

\hline
        
%%\cline{3-8}

\multirow{4}{6em}{Transformers} 
& Small BERT + CNN   & 0.89         & 0.90          & 0.90           & 0.92          & 0.92          & 0.92          \\

& Small BERT + MLP   & 0.90          & 0.90          & 0.90           & 0.92          & 0.92          & 0.92          \\

& ELECTRA + CNN  & 0.93         & 0.93         & 0.93          & 0.94          & 0.94          & 0.94          \\

& ELECTRA + MLP   & 0.95         & 0.93         & \textbf{0.94}          & 0.96          & 0.96          & \textbf{0.95}          \\

\hline
\multicolumn{8}{c}{\textbf{Testing - Founta}}                                                                                                                                           \\
\hline

\multirow{2}{*}{Glove}              
& Glove + CNN                                         & 0.55         & 0.54         & \textbf{0.54}          & 0.57          & 0.59          & \textbf{0.57}          \\
& Glove + MLP                                         & 0.50          & 0.50          & 0.49          & 0.53          & 0.57          & 0.54          \\
\hline

\multirow{4}{*}{Transformers}       

& Small BERT + CNN                                    & 0.51         & 0.51         & 0.50           & 0.54          & 0.55          & 0.54          \\
& Small BERT + MLP                                    & 0.51         & 0.51         & 0.50           & 0.54          & 0.55          & 0.54          \\
& ELECTRA + CNN                                       & 0.49         & 0.49         & 0.49          & 0.52          & 0.55          & 0.53          \\
& ELECTRA + MLP                                       & 0.49         & 0.49         & 0.49          & 0.52          & 0.55          & 0.53          \\
                                    \hline
                                    
\multicolumn{8}{c}{\textbf{Testing - TSA}}    \\

\hline
\multirow{2}{*}{Glove}              
& Glove + CNN                                         & 0.49         & 0.46         & 0.46          & 0.87          & 0.73          & 0.79          \\
& Glove + MLP                                         & 0.50          & 0.49         & 0.46          & 0.87          & 0.70           & 0.77          \\

\hline
\multirow{4}{*}{Transformers}       

& Small BERT + CNN                                    & 0.55         & 0.60          & 0.56          & 0.89          & 0.83          & 0.85          \\
& Small BERT + MLP                                    & 0.55         & 0.59         & 0.56          & 0.89          & 0.85          & 0.87          \\
& ELECTRA + CNN                                       & 0.61         & 0.56         & \textbf{0.58}          & 0.89          & 0.91          & \textbf{0.90}           \\
& ELECTRA + MLP                                       & 0.63         & 0.56         & \textbf{0.58}          & 0.89          & 0.92          & \textbf{0.90}           \\
                                    \hline
                                    \hline

%%\cline{3-8}
\end{tabular}

\end{table}

\textbf{Results in Scenario 2}. In the experiment scenario 2, we train models on the Founta dataset and evaluate them on the other two datasets, with the results on the Founta test set as the baseline. The results are presented in Table \ref{crossd-Founta}. The models generalize poorly from Founta to Davidson, which is similar to the results of generalizing from Davidson to Founta. However, it is interesting that when we test the models on Davidson dataset, transformer embeddings used with MLP-based models enable really good results, with macro average $F_1$ leading to 0.81 and weighted average $F_1$ leading to 0.85.

On the Founta Dataset, all models achieve excellent performance since there is no domain shift in this case. Here the transformer embeddings provide better results than Glove models which reach 0.93 in macro average and 0.93-0.94 in a weighted average for ELECTRA and Small BERT. Glove models also achieve good results with $F_1$ score of 0.90-0.91.
% but still lagging behind the transformers embedding.

On the experiments on the TSA dataset, transformer embeddings obtain well performance reaching to 0.49-0.50 on Precison for Small BERT and ELECTRA for macro average. For weighted average, the performance is noted by 0.86 for ELECTRA, Outperforming Glove models with 0.68 and 0.75.

\begin{table}[h!]
\centering
%%\resizebox{\textwidth}{!}
% \fontsize{8}{12}\selectfont

\caption{Scenario 2 - Cross-domain experimental results with Founta as the source domain.}
\label{crossd-Founta}

\begin{tabular}{llllllllll}
\hline

\multirow{2}{3em}{\textbf{Embedding}} &\multirow{2}{5em}{\textbf{Model}} & \multicolumn{3}{ c }{\textbf{Macro}} & \multicolumn{3}{ c }{\textbf{Weighted Avg.}}  \\ 

% \cline{3-8}

&& \textbf{P} & \textbf{R} & \textbf{$F_1$} & \textbf{P}&\textbf{R}&\textbf{$F_1$}  \\
\hline
\hline

% \multicolumn{8}{c}{\textbf{Training/ Validation - Founta}}                                                                                                                              \\
% \hline
% \hline
\multicolumn{8}{c}{\textbf{Testing - Davidson}}                                                                                                                                         \\
\hline
\hline

\multirow{2}{*}{Glove}              
& Glove + CNN & 0.55&0.56&0.55&0.66&0.60&0.62 \\
& Glove + MLP   & 0.49&0.49&0.48&0.6&0.58&0.59            \\

\hline
\multirow{4}{*}{Transformers}       

& Small BERT + CNN & 0.76&0.80&0.77&0.83&0.81&0.82 \\
& Small BERT + MLP & 0.79&0.83& \textbf{0.81}&0.86&0.84& \textbf{0.85} \\

& ELECTRA + CNN & 0.80&0.83&0.80&0.86&0.85& \textbf{0.85} \\
& ELECTRA + MLP   &0.80&0.85&\textbf{0.81}&0.87&0.84& \textbf{0.85} \\
                                    
\hline
\multicolumn{8}{c}{\textbf{Testing - Founta}}  \\
\hline
\hline
\multirow{2}{*}{Glove}              
& Glove + CNN  &0.91&0.91&0.91&0.92&0.92&0.92 \\
& Glove + MLP  &0.90&0.90&0.90&0.91&0.91&0.91 \\
\hline
\multirow{4}{*}{Transformers}       
& Small BERT + CNN  &0.93&0.93&\textbf{0.93}&0.93&0.93&0.93 \\
& Small BERT + MLP  &0.93&0.93& \textbf{0.93}&0.93&0.93&0.93 \\

& ELECTRA + CNN &0.94&0.93&\textbf{0.93}&0.94&0.94& \textbf{0.94}  \\
& ELECTRA + MLP  &0.93&0.93&\textbf{0.93}&0.94&0.94& \textbf{0.94}  \\
                                    
\hline
\multicolumn{8}{c}{\textbf{Testing - TSA}}                                                                                                                                              \\
\hline
\hline
\multirow{2}{*}{Glove}              
& Glove + CNN &0.50&0.50&0.42&0.88&0.57&0.68 \\
& Glove + MLP &0.49&0.48&0.45&0.87&0.67&0.75 \\

\hline
\multirow{4}{*}{Transformers}       
& Small BERT + CNN &0.51&0.51&0.49&0.87&0.78&0.82\\
& Small BERT + MLP &0.50&0.51&0.49&0.87&0.76&0.81 \\

& ELECTRA + CNN   &0.50&0.50 & \textbf{0.50} &0.87&0.85& \textbf{0.86} \\
& ELECTRA + MLP   &0.50&0.50 & \textbf{0.50} &0.87&0.86& \textbf{0.86} \\

\hline

%%\cline{3-8}
\end{tabular}

\end{table}

\textbf{Results in Scenario 3}. Here the models are trained on the TSA dataset and evaluated on the other two datasets. The results are shown in Table \ref{crossd-TSA}.
% The pattern in result are quite similar to what we have obtained in scenario 1. 

When the testing is performed on the Davidson dataset, transformer embeddings are better than Glove models. Small BERT with CNN can obtain a macro $F_1$ score of 0.61, followed by its combination with MLP at 0.60. 
% ELECTRA provided the accraucies of 0.57 and 0.55 with CNN and MLP.
Glove models obtain a precision of  0.45/0.43. In terms of weighted average score, Small BERT with CNN is still the best performer, achieving a weighted average $F_1$ score of 0.73, followed by Small BERT with MLP that obtains a score of 0.72. ELECTRA is the second best with $F_1$ score ranging in 0.70 and 0.71. Glove lags behind with $F_1$ score at 0.61/0.62.

In case of the Founta data, the pattern in prediction is similar to that in Scenario 1. The best performance here is only 0.47 in macro $F_1$ score and 0.54 in weighted average $F_1$ score, both of which are achieved by Small BERT-based MLP models. Glove-based models obtain comparative performance to the transformers-based models in this case.
% It was noticed that in terms of w=macro average the glove with CNN and SmallBERT with MLP provided the same accuracy of 0.47 while the SmallBERT with CNN came as the second best model with accuracy at 0.45. ELECTRA and Glovw with MLP shared almost similar accuracy between 0.42 and 0.43. The weighted average accuracy was recorded to be highest with Small BERT with MLP followed by CNN at 0.54 and 0.53. Glove and ELECTRA were almost similar with accuracy going lowest with ELECTRA with MLP at 0.50 to highest recorded with glove with CNN at 0.52.

When the testing is performed on TSA dataset, the results get much better across the models, as they are trained on the TSA training data. For the macro average measure, the highest performance is shared between Small BERT and ELECTRA with $F_1$ score ranging between 0.86 to 0.88. Glove provides the best performance with an $F_1$ score of 0.75 in combination with MLP. For the weighted average measure, the highest performance is also achieved by transformer embeddings-based models, with the averaged $F_1$ score  between 0.96-0.97. Glove models come next with the $F_1$ score of 0.90 and 0.94 with CNN and MLP, respectively.

\begin{table}[h!]
\centering
%%\resizebox{\textwidth}{!}
% \fontsize{8}{12}\selectfont

\caption{Scenario 3 - Cross-domain experimental results with TSA as the source domain.}
\label{crossd-TSA}

\begin{tabular}{llllllllll}
\hline

\multirow{2}{3em}{\textbf{Embedding}} &\multirow{2}{5em}{\textbf{Model}} & \multicolumn{3}{ c }{\textbf{Macro}} & \multicolumn{3}{ c }{\textbf{Weighted Avg.}}  \\ 

% \cline{3-8}

&& \textbf{P} & \textbf{R} & \textbf{$F_1$} & \textbf{P}&\textbf{R}&\textbf{$F_1$}  \\
\hline
\hline

% \multicolumn{8}{c}{\textbf{Training/ Validation - TSA}}                                                                                                                                 \\
% \hline
\multicolumn{8}{c}{\textbf{Testing - Davidson}}                                                                                                                                         \\
\hline
\multirow{2}{*}{Glove}              
& Glove + CNN
& 0.45         & 0.48         & 0.45          & 0.58          & 0.67          & 0.61          \\
                                    
& Glove + MLP
& 0.45         & 0.49         & 0.43          & 0.58          & 0.71          & 0.62          \\
                                    
\hline                                    
\multirow{4}{*}{Transformers}       
& Small BERT + CNN                                    
& 0.70          & 0.60          & \textbf{0.61}          & 0.74          & 0.77          & \textbf{0.73}          \\

& Small BERT + MLP                                    
& 0.67         & 0.59         & 0.60           & 0.72          & 0.76          & 0.72          \\

& ELECTRA + CNN                                       
& 0.67         & 0.57         & 0.57          & 0.72          & 0.76          & 0.71          \\

& ELECTRA + MLP                                       
& 0.75         & 0.56         & 0.55          & 0.76          & 0.77          & 0.70           \\
                                    \hline
\multicolumn{8}{c}{\textbf{Testing - Founta}}                                                                                                                                           \\
\hline
\multirow{2}{*}{Glove}              
& Glove + CNN                                         
& 0.48         & 0.49         & \textbf{0.47}          & 0.51          & 0.55          & 0.52          \\

& Glove + MLP                                        
& 0.52         & 0.50          & 0.43          & 0.55          & 0.61          & 0.51          \\
\hline                                    
                                    
\multirow{4}{*}{Transformers}       

& Small BERT + CNN                                    
& 0.53         & 0.51         & 0.45          & 0.55          & 0.61          & 0.53          \\
& Small BERT + MLP                                    
& 0.53         & 0.51         & \textbf{0.47}          & 0.55          & 0.51          & \textbf{0.54}          \\
& ELECTRA + CNN                                       
& 0.50          & 0.50          & 0.43          & 0.53          & 0.61          & 0.51          \\
& ELECTRA + MLP               
& 0.51         & 0.50          & 0.42          & 0.54          & 0.62          & 0.50           \\
  \hline

\multicolumn{8}{c}{\textbf{Testing - TSA}}                                                                                                                                              \\
\hline
\multirow{2}{*}{Glove}              
& Glove + CNN
& 0.66         & 0.84         & 0.71          & 0.94          & 0.88          & 0.90           \\
                                    
& Glove + MLP
& 0.73         & 0.78         & 0.75          & 0.94          & 0.93          & 0.94          \\

\hline

\multirow{4}{*}{Transformers}       
& Small BERT + CNN                                    
& 0.85         & 0.86         & 0.86          & 0.96          & 0.96          & 0.96          \\

& Small BERT + MLP
& 0.86         & 0.88         & 0.87          & 0.97          & 0.96          & 0.96          \\
                                    
& ELECTRA + CNN
& 0.88         & 0.87         & \textbf{0.88}          & 0.97          & 0.97          & \textbf{0.97}          \\

& ELECTRA + MLP
& 0.91         & 0.85         & \textbf{0.88}          & 0.97          & 0.97          & \textbf{0.97}   \\
                                    \hline
%%\cline{3-8}
\end{tabular}

\end{table}

% \subsubsection{Analysis}

\textbf{Overall Observations}. The transformers-based methods generally perform better than the Glove-based methods. The transformers enable the downstream detectors to obtain  good performance in detecting non-hatred tweets across all three settings (as indicated by the performance in weighted average $F_1$ score) and the impressive performance on detecting hatred tweets in some scenarios. It is interesting that the cross-domain performance is not reversible in our results. For example, transformers-based methods work well when transferring knowledge from the Founta data to the Davidson data in Table \ref{crossd-Founta}, but they work poorly in the reverse case in Table \ref{crossd-wzls}. This may be because the Founta data is much larger than the Davidson data, containing knowledge of a broader hate speech scope (\ie, hate and abusive tweets) than Davidson (\ie, hate class only). As a result, the Founta data may contain hate speech knowledge relevant to that of Davidson, but the hatred tweets in the Davidson data may be not generalizable to that in Founta. This explains the different result patterns on Davidson and Founta in Tables \ref{crossd-wzls} and \ref{crossd-Founta}.

TSA explicitly focuses on racist and sexism tweets, which is different from the other two datasets. As a result, it is difficult to adapt relevant knowledge from the other two datasets to the hate speech detection on TSA, as shown in Tables \ref{crossd-wzls} and \ref{crossd-Founta}. Further, the models including the transformers-based methods, also fail to adapt relevant knowledge from TSA to Davidson and Founta, indicating nearly no shared knowledge of hatred tweets between TSA and the other two datasets.

\section{Conclusions and Discussions}\label{sec:conclusions}

% We have used our research to answer mainly five question which are described in section \ref{accuracy} to \ref{cde}. We implemented in total fourteen models for hate speech detection to give a comparison which model is more suitable for the purpose. We included three models with TF-IDF, three models with Glove and eight models with Transformers. We have tried to perform the experiment by taking complete dataset and also by modifying the dataset to cover cross domain for detection of hate speech. 

This paper presents a large-scale empirical evaluation of 14 shallow and deep models for hate speech detection on three commonly-used benchmarks of different data characteristics. This is to provide important insights into their detection effectiveness, computational efficiency, capability in using pre-trained models, and domain generalizability for their deployment in real-world applications. Our conclusions based on the empirical results above are as following.
\begin{itemize}
    \item \textbf{Detection effectiveness}. As shown in Table \ref{table:2}, the combination of BERT, ELECTRA, and Al-BERT and neural network-based classifiers perform consistently better than the other methods on the three benchmarks, especially in macro $F_1$ score. It is interesting that TF-IDF-enabled shallow classifiers (\eg, TF-IDF + XGB) can outperform Glove-enabled deep classifiers on most datasets.
    
    \item \textbf{Computational efficiency}. In terms of computational cost, transformers-enabled classifiers are more costly than the other models, as illustrated in Figure \ref{ct}. AI-BERT is the most time-consuming model among the transformers, followed by the BERT model. Small BERT is the most efficient transformer model in our task. Deep classifiers like Bi-LSTM are also computationally costly. When considering both effectiveness and efficiency, ELECTRA-based MLP models seem to be the most practical method, which achieves among the best classification performance while being sufficiently computationally efficient.
    \item \textbf{Capability in using pre-trained models}. As transformers are pre-trained on significantly larger corpora than Glove, they learn embedding space with richer semantics, empowering significantly better detection performance. This can be observed by the performance difference of having the same MLP-based classifier trained upon respective TF-IDF, Glove, and transformers-based representations, as shown in Table \ref{table:2}. However, it does not mean that the larger the pre-trained models are, the better performance the model would obtain. For example, the ELECTRA transformer can normally perform better than, or on par with, the larger transformer BERT on a number of cases in Table \ref{table:1}.
    \item \textbf{Domain generalizability}.  Benefiting from the large-scale pre-training, \ref{crossd-Founta} and \ref{crossd-TSA}, both Glove and transformers learn generalizable pre-trained embeddings, as shown in Tables \ref{crossd-wzls}. Again, due to the larger corpora used and the greater model capacity, the transformers-enabled classifiers gain stronger cross-domain generalization ability than the same Glove-based classifiers. It is interesting that the cross-domain performance may be not reversible due to the difference in the data size and the types of hatred classes covered in the source and target domains.
    
\end{itemize}
    
Although the transformers-based hate speech detectors show promising performance, they are still weak in terms of macro $F_1$ performance (\eg, the best performer gains a macro $F_1$ of 0.68 on the Founta dataset and 0.90 on the TSA dataset), indicating the great difficulty in achieving high precision and recall rates of identifying the minority hatred tweets from the massive tweets. Another major challenge lies in domain generalization ability is that it is difficult to collect a dataset that well covers all possible properties of hate speech in social media, which may lead to a domain difference between the source data and the target data. Thus, it is important for detectors to have a good domain generalization ability in practice. However, as shown in our results, there is still a large gap between the same-domain performance and the cross-domain performance. These two challenges present some important opportunities to further promote the development and deployment of hate speech detection techniques.

\appendix
% \newpage
% \section*{Declarations}

% \noindent\textbf{Ethical approval} - 
% Not applicable.\\
 
% \noindent\textbf{Competing interests} - 
% Not applicable.\\

% \noindent\textbf{Authors' contributions} - 
% \textbf{Hezhe Qiao}: Methodology, Investigation, Writing- Original draft preparation. \textbf{Jitendra Singh Malik}: Methodology, Investigation, Validation, Writing- Original draft preparation. \textbf{Guansong Pang}: Conceptualization, Methodology, Writing- Reviewing and Editing, Supervision. \textbf{Anton van den Hengel}: Resources, Writing- Reviewing and Editing.\\
 
% \noindent\textbf{Funding} - 
% Not applicable.\\
 
% \noindent\textbf{Availability of data and materials} - 
% Code and data are available at \href{https://github.com/jmjmalik22/Hate-Speech-Detection}{GitHub}.\\

\bibliographystyle{plain}
\bibliography{mybib}

\end{document}